\documentclass[11pt,a4paper,usenames,dvipsnames]{article}
\usepackage[hyperref]{acl2019}
\usepackage{times}
\usepackage{latexsym}
\usepackage{url}
\usepackage{xcolor}
\usepackage{multirow}
\usepackage{graphicx}
\usepackage{booktabs}
\usepackage{amsmath}

\usepackage{tikz}
\usetikzlibrary{calc}
\usetikzlibrary{decorations.pathmorphing}
\usetikzlibrary{shapes,arrows,chains}
\usetikzlibrary{shapes.geometric}
\usetikzlibrary{patterns}
\usetikzlibrary{positioning}
\tikzset{>=latex}

\newcommand{\gn}[1]{\textcolor{magenta}{\bf\small [#1 --GN]}}

\newcommand{\yhl}[1]{\textcolor{Green}{\bf\small [#1 --YHL]}}

\newcommand{\langrankns}{\textsc{LangRank}}
\newcommand{\langrank}{\langrankns~}

\newcommand\blfootnote[1]{%
  \begingroup
  \renewcommand\thefootnote{}\footnote{#1}%
  \addtocounter{footnote}{-1}%
  \endgroup
}

\aclfinalcopy % Uncomment this line for the final submission
\def\aclpaperid{1846} %  Enter the acl Paper ID here

\title{Choosing Transfer Languages for Cross-Lingual Learning}

\author{Yu-Hsiang Lin$^{*}$, Chian-Yu Chen$^{*}$, Jean Lee$^{*}$, Zirui Li$^{*}$, Yuyan Zhang${^*}$, \\ 
\textbf{Mengzhou Xia, Shruti Rijhwani, Junxian He, Zhisong Zhang, Xuezhe Ma, } 
\\ \textbf{Antonios Anastasopoulos, Patrick Littell$^{\dag}$, Graham Neubig} \\
  Language Technologies Institute, Carnegie Mellon University \\
  $^\dag$National Research Council, Canada}

\date{}

\begin{document}
\maketitle

\begin{abstract}
Cross-lingual transfer, where a high-resource \emph{transfer} language is used to improve the accuracy of a low-resource \emph{task} language, is now an invaluable tool for improving performance of natural language processing (NLP) on low-resource languages.
However, given a particular task language, it is not clear \emph{which} language to transfer from, and the standard strategy is to select languages based on \emph{ad hoc} criteria, usually the intuition of the experimenter.
Since a large number of features contribute to the success of cross-lingual transfer (including phylogenetic similarity, typological properties, lexical overlap, or size of available data), even the most enlightened experimenter rarely considers all these factors for the particular task at hand.
In this paper, we consider this task of automatically selecting optimal transfer languages as a ranking problem, and build models that consider the aforementioned features to perform this prediction.
In experiments on representative NLP tasks, we demonstrate that our model predicts good transfer languages much better than \emph{ad hoc} baselines considering single features in isolation, and glean insights on what features are most informative for each different NLP tasks, which may inform future \emph{ad hoc} selection even without use of our method.%
\blfootnote{\hspace{-1.5mm}$^{*}$Equal contribution}%
\footnote{Code, data, and pre-trained models are available at \url{https://github.com/neulab/langrank}}

% We examine the strategies of selecting the best languages to be the source of cross lingual transfer for a given NLP task in a low-resource language. Different from the  commonly used approaches that

% Trained on 2,862 transfer pairs for machine translation, 72 pairs for entity linking, and 4,260 pairs for POS tagging, we found that the languages selected by the ranking models are substantially better than those selected based on any single language or corpus attribute, improving NDCG@3 by 17\%, 12\%, and XX\% for machine translation, entity linking, and POS tagging, respectively. \yhl{TODO: Add dependency parsing later.}

\end{abstract}

% ----------------------------------------------------
% ----------------------------------------------------

\section{Introduction}
\label{sec:intro}

%\gn{Overall, try to make all of your section titles be in title case.}

%\gn{Once you've introduced an acronym like ``MT'', make sure you use it for the rest of the paper and don't sometimes use it and sometimes spell out like ``machine translation''.}

% Notes
% For introduction: one paragraph for each; for abstract, one sentence for each
% 1) Background: the desire to improve the NLP task for low-resource. It is challenging to ... for low-resource language. Cross lingual has proven effective to solve.. but the question is which language to transfer from
% 2) Existing approaches and problems: it is not clear how to choose the language to perform the transfer learning. (based on heuristics/intuition)
% 3) What do we do
% 4) What are the results: how much better we are doing compared to traditional approach
% 5) Use cases: 1. take insights without actually training a model (datasize and similarity are both important) 2. Release the pre-trained model, release model for 4 different type of tasks (so people can pick the model that is the closet to the task they are doing)

% \gn{This is nicely colored! If the person who made this figure can send me an editable file I can squeeze it to make it use less space.}

\begin{figure}[!t]
\centering
\includegraphics[width=\columnwidth]{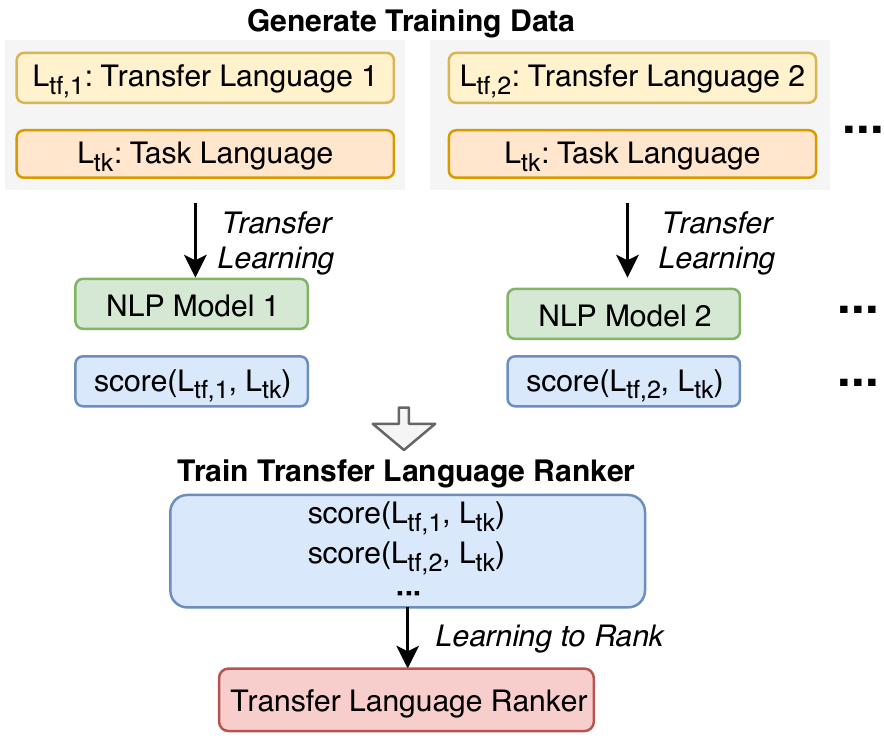}
\caption{Workflow of learning to select the transfer languages for an NLP task: (1) train a set of NLP models with all available transfer languages and collect evaluation scores, (2) train a ranking model to predict the top transfer languages.}
\label{fig:flowchart} 
% \vspace{-4mm}
\end{figure}

A common challenge in applying natural language processing (NLP) techniques to low-resource languages is the lack of training data in the languages in question.
% This is especially a hurdle to documenting endangered and other very-low-resource languages.
It has been demonstrated that through cross-lingual transfer, it is possible to leverage one or more similar high-resource languages to improve the performance on the low-resource languages in several NLP tasks, including machine translation \cite{zoph2016transfernmt, johnson17googlemultilingual, nguyen-chiang:2017:I17-2, Neubig2018}, parsing \cite{tackstrom12crosslingualsyntax, ammar16manylanguages, ahmad2018near, ponti2018isomorphic}, part-of-speech or morphological tagging \cite{tackstrom2013token,cotterell17crosslingualmorphology,malaviya2018factorgraph,plank2018disparate}, named entity recognition \cite{zhang2016nerannotationprojection,mayhew2017cheap,xie2018crosslingual}, and entity linking \cite{tsai2016crosslingualwikification,rijhwani19aaai}.
There are many methods for performing this transfer, including joint training \cite{ammar16manylanguages,tsai2016crosslingualwikification,cotterell17crosslingualmorphology,johnson17googlemultilingual,malaviya2018factorgraph}, annotation projection \cite{tackstrom12crosslingualsyntax,tackstrom2013token,zhang2016nerannotationprojection,ponti2018isomorphic,plank2018disparate}, fine-tuning \cite{zoph2016transfernmt,Neubig2018}, data augmentation \cite{mayhew2017cheap}, or zero-shot transfer \cite{ahmad2018near,xie2018crosslingual,Neubig2018,rijhwani19aaai}. The common thread is that data in a high-resource \emph{transfer language} is used to improve performance on a low-resource \emph{task language}.

However, determining the best transfer language for any particular task language remains an open question -- the choice of transfer language has traditionally been done in a heuristic manner, often based on the intuition of the experimenter.
A common method of choosing transfer languages involves selecting one that belongs to the same language family or has a small phylogenetic distance in the language family tree to the task language~\cite{dong15multitask, johnson17googlemultilingual, cotterell17crosslingualmorphology}.
However, it is not always true that all languages in a single language family share the same linguistic properties~\cite{ahmad2018near}. Therefore, another strategy is to select transfer languages based on the typological properties that are relevant to the specific NLP task, such as word ordering for parsing tasks~\cite{ammar16manylanguages, ahmad2018near}.
With several heuristics available for selecting a transfer language, it is unclear \emph{a priori} if any single attribute of a language will be the most reliable criterion in determining whether cross-lingual learning is likely to work for a specific NLP task. Other factors, such as lexical overlap between the training datasets or size of available data in the transfer language, could also play a role in selecting an appropriate transfer language.
% \sr{if these factors haven't been used in previous work, but show positive results with our method, we can state that here or in the analysis section}
% The answers to these questions may also be task-dependent; the criteria may change from one NLP task to another.
Having an empirical principle regarding how to choose the most promising languages or corpora to transfer from has the potential to greatly reduce the time and effort required to find, obtain, and prepare corpora for a particular language pair.

In this paper, we propose a framework, which we call \langrankns, to empirically answer the question posed above: \emph{given a particular task low-resource language and NLP task, how can we determine which languages we should be performing transfer from?}
We consider this language prediction task as a ranking problem, where each potential transfer language is represented by a set of attributes including typological information and corpus statistics, such as word overlap and dataset size.
Given a task language and a set of candidate transfer languages, the model is trained to rank the transfer languages according to the performance achieved when they are used in training a model to process the task low-resource language.
These models are trained by performing a computation- and resource-intensive exhaustive search through the space of potential transfer languages, but at test time they can rapidly predict optimal transfer languages, based only on a few dataset and linguistic features, which are easily obtained.

% without performing explicit model training or even resource acquisition. \sr{What resources are we talking about here? If we're trying to say that it's extremely fast/easy to get the ranks at test time, let's say that explicitly.}

In experiments, we examine cross-lingual transfer in four NLP tasks: machine translation (MT), entity linking (EL), part-of-speech (POS) tagging and dependency parsing (DEP).
We train gradient boosted decision trees (GBDT; \newcite{Ke2017}) to select the best transfer languages based on the aforementioned features.
% \an{the features are not mentioned above}
% For machine translation, we used 54 languages in the TED corpus \cite{Qi2018}, generating 2,862 transfer pairs. For entity linking \yhl{What dataset?}, we generated 72 transfer pairs. For POS tagging and dependency parsing \yhl{What dataset?}, we generated 4,260 pairs.
We compare our ranking models with several reasonable baselines inspired by the heuristic approaches used in previous work, and show that our ranking models significantly improve the quality of the selection of the top languages for cross lingual transfer.
% Using our ranking model, the normalized discounted cumulative gain (NDCG) of selecting the top 3 transfer languages for machine translation, entity linking, and POS tagging is improved by 17\%, 12\%, and 8\% against the best baseline methods, respectively. \gn{I commented this out, because I'm not sure if NDCG is such an intuitive measure.}
In addition, through an ablation study and examining the learned decisions trees, we glean insights about which features were found to be useful when choosing transfer languages for each task. This may inform future attempts for heuristic selection of transfer languages, even in the absence of direct use of \langrankns. 

% ----------------------------------------------------
% ----------------------------------------------------

\section{Problem Formulation}
\label{sec:Formulation}

We define the \emph{task} language $t$ as the language of interest for a particular NLP task, and the \emph{transfer} language $a$ as the additional language that is used to aid in training models.
Formally, during the training stage of transfer learning, we perform a model training step:
\begin{equation*}
  M_{t,a} = \text{train}(\langle x_t^{(trn)}, y_t^{(trn)} \rangle, \langle x_a^{(trn)}, y_a^{(trn)} \rangle),
\end{equation*}
where $x^{(trn)}$ and $y^{(trn)}$ indicate input and output training data for each training language, and $M_{t,a}$ indicates the resulting model trained on languages $t$ and $a$.
The actual model and training procedure will vary from task to task, and we give several disparate examples in our experiments in \S\ref{sec:tasks}.
The model can then be evaluated by using it to predict outputs over the test set, and evaluating the results:
\begin{align*}
  \hat{y}_{t,a}^{(tst)} & = \text{predict}(x_t^{(tst)}; M_{t,a}) \\
  c_{t,a} & = \text{evaluate}(y_t^{(tst)}, \hat{y}_{t,a}^{(tst)}),
\end{align*}
where $c_{t,a}$ is the resulting test-set score achieved by using $a$ as an transfer language. 

Assuming we want to get the highest possible performance on task language $t$, one way to do so is to exhaustively enumerate over every single potential transfer language $a$, train models, and evaluate the test set.
In this case, the optimal transfer language for task language $t$ can be defined as:
\begin{equation*}
  a_t^{*} = \text{argmax}_{a} c_{t,a}.
\end{equation*}
However, as noted in the introduction, this brute-force method for finding optimal transfer languages is not practical: if resources for many languages are available \emph{a priori}, it is computationally expensive to train all of the models, and in many cases these resources are not-available \emph{a priori} and need to be gathered from various sources before even starting experimentation.

Thus, we turn to formulating our goal as a ranking task: given an NLP task, a low-resource task language $t$, and a list of $J$ available high-resource transfer languages $a_1, a_2, \ldots, a_J$, attempt to predict their ranking according to their expected scores $c_{t,a_1}, c_{t,a_2}, \ldots, c_{t,a_J}$ without actually calculating the scores themselves. 
To learn this ranker, we need to first create training data for the ranker, which we create by doing an exhaustive sweep over a set of \emph{training} task languages $t_1, t_2, \ldots, t_I$, which results in sets of scores $\{c_{t_1,a_1},\ldots,c_{t_1,a_J},\},\ldots,\{c_{t_I,a_1},\ldots,c_{t_I,a_J}\}$.
These scores that can be used to train a ranking system, using standard methods for learning to rank (see, e.g., \newcite{liu2009learning}).
Specifically, these methods work by extracting features from the pair of languages $\langle t_i,a_j \rangle$:
\begin{equation*}
  \phi_{t_i,a_j} = \text{feat\_extract}(t_i, a_j)
\end{equation*}
and then using these features to predict a relative score for each pair of task and transfer languages
\begin{equation*}
  r_{t_i,a_j} = \text{rank\_score}(\phi_{t_i,a_j}; \theta)
\end{equation*}
where $\theta$ are the parameters of the ranking model.
These parameters $\theta$ are learned in a way such that the order of the ranking scores $r_{t_i,a_1},\ldots,r_{t_i,a_J}$ match as closely as possible with those of the gold-standard evaluation scores $c_{t_i,a_1},\ldots,c_{t_i,a_J}$.

Now that we have described the overall formulation of the problem, there are two main questions left: how do we define our features $\phi_{t_i,a_j}$, and how do we learn the parameters $\theta$ of the ranking model?

\section{Ranking Features}
\label{sec:features}

%\gn{Many of the variable names here use full words. In general, we prefer variable names to be single characters (see my writing in the previous section). Alternatively, you can have functions that are spelled out. Please try to match this (maybe define these long names as functions instead?). Also, see the following guide for math typesetting: \url{http://demo.clab.cs.cmu.edu/cdyer/short-guide-typesetting.pdf}} \yhl{Revised.}

% \sr{re-iterate that some of these are based on previous work, so it doesn't look like we're using arbitrary attributes}

We represent each language pair/corpus by a set of features, split into two classes: dataset-dependent and dataset-independent.

\subsection{Data-dependent Features}

Dataset-dependent features are statistical features of the particular corpus used, such as dataset size and the word overlap between two corpora.
Importantly, these features require the dataset to already be available for processing and thus are less conducive to use in situations where resources have not yet been acquired.
Specifically, we examine the following categories:

\paragraph{Dataset Size:}
We denote the number of training examples in the transfer and task languages by $s_{tf}$ and $s_{tk}$, respectively.
%\sr{can we use $t$ and $a$ as introduced in the previous section instead of $tf$ and $tk$?} \sr{$s$ is used to represent the scores in the previous section}.
For MT, POS and DEP, this is the number of sentences in a corpus, and for EL the dataset size is the number of named entities in a bilingual entity gazetteer.
In our experiments, we also consider the ratio of the dataset size, $s_{tf} / s_{tk}$, as a feature, since we are interested in how much bigger the transfer-language corpus is than the task-language corpus.

%For POS, the dataset size is the sum of number of sentences in training, validation, and test sets, if available; some low-resource languages 
%we combine training, validation, and test sets and compute the number of sentences in the combined corpus as we do not have enough data for low-resource task languages to form training set in POS tagging task.

\paragraph{Type-Token Ratio (TTR):}
The TTR of the transfer- and task-language corpora, $t_{tf}$ and $t_{tk}$, respectively, is the ratio between the number of types (the number of unique words) and the number of tokens \cite{richards87}. It is a measure for lexical diversity, as a higher TTR represents higher lexical variation. We also consider the distance between the TTRs of the transfer- and task-language corpora, which may very roughly indicate their morphological similarity:
{\setlength{\abovedisplayskip}{5pt}
\setlength{\belowdisplayskip}{5pt}
\[
    d_{ttr} = \left( 1 - \frac{t_{tf}}{t_{tk}} \right)^2.
\]
}
Transfer and task languages that have similar lexical diversity are expected to have $d_{ttr}$ close to 0.

The data for the entity linking task consists only of named entities, so the TTR is typically close to 1 for all languages. Therefore, we do not include TTR related features for the EL task.

% We compute $r_{tf}$, $t_{tk}$ and $d_{ttr}$ from the training set of the machine translation task and from both the training and testing set of the POS tagging task.

\paragraph{Word Overlap and Subword Overlap:}
We measure the similarity between the vocabularies of task- and transfer-language corpora by word overlap $o_{w}$, and subword overlap $o_{sw}$:
\begin{align*}
    o_{w} &= \frac{|T_{tf} \cap T_{tk}|}{|T_{tf}| + |T_{tk}|},
    &o_{sw} = \frac{|S_{tf} \cap S_{tk}|}{|S_{tf}| + |S_{tk}|},
\end{align*}
where $T_{tf}$ and $T_{tk}$ are the sets of types in the transfer- and task-language corpora, and $S_{tf}$ and $S_{tk}$ are their sets of subwords. The subwords are obtained by an unsupervised word segmentation algorithm \cite{Sennrich2016, Kudo2018}. Note that for EL, we do not consider subword overlap, and the word overlap is simply the count of the named entities that have exactly the same representations in both transfer and task languages. We also omit subword overlap in the POS and DEP tasks, as some low-resource languages do not have enough data for properly extracting subwords in the corpora used for training the POS and DEP models in our experiments.

% To accommodate languages that do not use white spaces as the word divider, we compute sub-word overlap ${o_{sw}}$ \cite{Sennrich2016, Kudo2018} between the transfer and task language corpus. We define $o_{sw}$ as: $\frac{Subwords_{tf} \cap Subwords_{tk}}{Tokens_{tf} + Tokens_{tk}}$, where $Subwords_{tf}$ and $Subwords_{tk}$ are the number of sub-word units in the transfer language corpus and task language corpus, respectively. 
% To obtain the  compute the number of subword units, we use an unsupervised word segmentation algorithm \cite{GoogleSentencePiece} to compute the number of sub-word units in the corpus. We compute sub-word overlap from the training set of the machine translation task. As the algorithm requires sufficient data to properly estimate the number of sub-word units, we omit this feature for the entity linking task and POS tagging task due to limited data.

\subsection{Dataset-independent Features}

Dataset-independent features are measures of the similarity between a pair of languages based on phylogenetic or typological properties established by linguistic study.
Specifically, we leverage six different linguistic distances queried from the URIEL Typological Database \cite{littell17}: 

\paragraph{Geographic distance ($d_{geo}$):}
The orthodromic distance between the languages on the surface of the earth, divided by the antipodal distance, based primarily on language location descriptions in Glottolog \cite{glottolog}.
\paragraph{Genetic distance ($d_{gen}$):}
The genealogical distance of the languages, derived from the  hypothesized tree of language descent in Glottolog.
\paragraph{Inventory distance ($d_{inv}$):}
The cosine distance between the phonological feature vectors derived from the PHOIBLE database \cite{phoible}, a collection of seven phonological databases. 
\paragraph{Syntactic distance ($d_{syn}$):}
The cosine distance between the feature vectors derived from the syntactic structures of the languages \cite{syntax}, derived mostly from the WALS database \cite{wals}.
\paragraph{Phonological distance ($d_{pho}$):}
The cosine distance between the phonological feature vectors derived from the WALS and Ethnologue databases \cite{lewis_2009}. 
\paragraph{Featural distance ($d_{fea}$):}
The cosine distance between feature vectors combining all 5 features mentioned above.

% ----------------------------------------------------

\section{Ranking Model}
\label{sec:ranking_model}

Having defined our features, the next question is what type of ranking model to use and how to learn its parameters $\theta$.
As defined in \S\ref{sec:Formulation}, the problem is a standard learning-to-rank problem, so there are a myriad of possibilities for models and learning algorithms \cite{liu2009learning}, and any of them would be equally applicable to our task.

We opt to use the GBDT \cite{Ke2017} model with LambdaRank as our training method \cite{Burges2010}.
This method works by learning an ensemble of decision-tree-based learners using gradient boosting, and specifically in our setting here has two major advantages.
First, its empirical performance -- it is currently one of the state-of-the-art methods for ranking, especially in settings that have few features and limited data.
Second, but perhaps more interesting, is its interpretability. Decision-tree based algorithms are relatively interpretable, as it is easy to visualize the learned tree structure.
One of our research goals is to understand what linguistic or statistical features of a dataset play important roles in transfer learning, so the interpretable nature of the tree-based model can provide valuable insights, which we elaborate further in \S\ref{sec:educatedguesses}.

\section{Experimental Settings}

% We outline the experimental setting in Figure \ref{flowchart}. We first perform transfer learning to obtain the performance score of each available transfer language. The results of the transfer learning are then used to train the ranker. \gn{Commented out because this is now described in Section 2.}

\subsection{Testbed Tasks}
\label{sec:tasks}

We investigate the performance of \langrank on four common NLP tasks: machine translation, entity linking, POS tagging, and dependency parsing.
We briefly outline the settings for all four NLP tasks, which are designed based on previous work on transferring between languages in these settings \cite{Neubig2018,rijhwani19aaai,kim-etal-2017-cross,ahmad2018near}.
% For each task, we first perform an exhaustive search through the space of all available transfer languages for the dataset at hand to obtain the true performance score of each transfer language for transfer learning. \gn{Commented out because this is }

%and will provide scripts with full experimental settings upon acceptance.

%\paragraph{Machine Translation}
\paragraph{Machine Translation}
We train a standard attention-based sequence-to-sequence model \cite{Bahdanau2015}, using the XNMT toolkit \cite{neubig18xnmt}.
We perform training on the multilingual TED talk corpus of \newcite{Qi2018}, using 54 task and 54 transfer languages, always translating into English, which results in 2,862 task/transfer pairs and 54 single-source training settings.
Transfer is performed by joint training over the concatenated task and transfer corpora, and subwords are learned over the concatenation of both corpora \cite{Sennrich2016}.

%\paragraph{Entity Linking}
\paragraph{Entity Linking}
The cross-lingual EL task involves linking a named entity mention in the task language to an English knowledge base. We train two character-level LSTM encoders, which are trained to maximize the cosine similarity between parallel (i.e., linked) entities~\cite{rijhwani19aaai}. We use the same dataset as~\newcite{rijhwani19aaai}, which contains language-linked Wikipedia article titles from 9 low-resource task languages and 53 potential transfer languages, resulting in 477 task/transfer pairs. We perform training in a zero-shot setting, where we train on corpora only in the transfer language, and test entity linking accuracy on the task language without joint training or fine-tuning.

\paragraph{POS Tagging}
We train a bi-directional LSTM-CNNs-CRF model~\citep{ma2016end} on word sequences without using pre-trained word embeddings. The implementation is based on the NCRF++ toolkit~\citep{yang2018ncrf}.
We perform training on the Universal Dependencies v2.2 dataset~\citep{ud22}, using 26 languages that have the least training data as task languages, and 60 transfer languages,\footnote{For each language, we choose the treebank that has the least number of training instances, which results in 60 languages with training data and 11 without training data.} resulting in 1,545 pairs of transfer-task languages.
Transfer is performed by joint training over the concatenated task and transfer corpora if the task language has training data, and training only with transfer corpora otherwise. The performance is measured by POS tagging accuracy on the task language.

\paragraph{Dependency Parsing}
For the dependency parsing task, we utilize a deep biaffine attentional graph-based model \cite{dozat2017deep}. We select 30 languages from Universal Dependencies v2.2 \cite{ud22}, resulting in 870 pairs of transfer-task languages. The selection basically follows the settings of \citet{ahmad2018near}, but we exclude Japanese (ja) since we observe unstable results on it. For this task, transfer is performed in the zero-shot setting where no task language annotations are available in training. We rely on the multi-lingual embeddings which are mapped into the same space with the offline method of \newcite{smith2017offline} and directly adopt the model trained with the transfer language to task languages. The performance is measured by LAS (Labeled Attachment Accuracy) excluding punctuation.
% \zs{0. please refer to `parsing.md' for some details, 1.wait for the parsing ranking results, 2. check consistence of the descriptions about parsing}

% ---------------------------
\iffalse
\yhl{I think the following statements are incorrect.}

For a specific task language, we obtain a list of performance scores $s_1$, $s_2$, \dots, $s_N$ associated with each available transfer languages $A_1$, $A_2$, ... , $A_N$. We define the rank $r_k$ of a transfer language $A_k$ as
\begin{align}
r_k &= \left\{
	\begin{array}{l}
	s_k - c + 1, \quad \textrm{if} \; s_k \geq c \\
	0, \quad\quad\quad\quad\;\;\; \textrm{otherwise}
	\end{array}
\right.
\end{align}
where $c$ is the cutoff. In our experiment, we set the cutoff to 10. By such setting, we train the model to only correctly rank top 10 transfer learning. 
\fi
% ------------------------------

\subsection{Evaluation Protocol}
\label{sec:eval_protocol}

We evaluate all our models on all NLP tasks with leave-one-out cross validation. For each cross-validation fold, we leave one language $\ell^{(tst)}$ out from the $N$ languages we have as the test set, and train our ranking model $\theta_{\ell^{(tst)}}$ using all remaining languages, $\{\ell^{(trn)}_1, \dots, \ell^{(trn)}_{N-1}\}$, as the training set.
During training, each $\ell^{(trn)}_i$ is treated as the task language in turn, and the other $N-2$ languages in the training set as transfer languages.
We then test the learned model $\theta_{\ell^{(tst)}}$ by taking $\ell^{(tst)}$ as the task language, and $\{\ell^{(trn)}_1, \dots, \ell^{(trn)}_{N-1}\}$ as the set of transfer languages, and predict the ranking scores $\{r_{\ell^{(tst)}, \ell^{(trn)}_1}, \dots, r_{\ell^{(tst)}, \ell^{(trn)}_{N-1}}\}$. We repeat this process with each language in all $N$ languages as the test language $\ell^{(tst)}$, and collect $N$ learned models.

%can then be used to produce a ranking $r_{t,\alpha_1},r_{t,\alpha_2},\ldots,r_{t,\alpha_J}$ of the available $J$ transfer languages.

%The produced ranking can be then compared to the gold-standard ranking that matches the order of the evaluation scores $s_{t,a_1},s_{t,a_2},\ldots,s_{t,a_J}$.
We use Normalized Discounted Cumulative Gain (NDCG) \cite{jarvelin2002cumulated} to evaluate the performance of the ranking model. The NDCG at position $p$ is defined as:
\begin{align*}
\ndcg@p = \frac{\dcg@p}{\idcg@p},
\end{align*}
where the Discounted Cumulative Gain (DCG) at position $p$ is
\begin{align*}
    \dcg@p = \sum_{i=1}^p \frac{2^{\gamma_i}-1}{\log_2(i+1)}.
\end{align*}
Here $\gamma_i$ is the relevance of the language ranked at position $i$ by the model being evaluated. We keep only the top-$\gamma_{\max}$ transfer languages as our learning signal: the true best transfer language has $\gamma = \gamma_{\max}$, and the second-best one has $\gamma = \gamma_{\max} - 1$, and so on until $\gamma = 1$, with the remaining languages below the top-$\gamma_{\max}$ ones all sharing $\gamma = 0$.
The Ideal Discounted Cumulative Gain (IDCG) uses the same formula as DCG, except it is calculated over the gold-standard ranking. When the predicted ranking matches the ``true" ranking, then NDCG is equal to 1.

%To compare the LambdaRank \sr{\langrank?} method with adhoc \sr{heuristic-based?} selection criteria, we also rank all transfer languages for each testing task language using single features. Specifically, we rank the transfer learning performance of each transfer language in the same order of the following features: the word-level overlap, subword-level overlap, transfer language data size, task language data size, as we consider the high overlap and data size will contribute more benefits to the transfer learning. In contrast, we rank the transfer learning performance of each transfer language in the reverse order of URIEL distances -- the smaller the distance, the closer the pair of languages are. In the result section, we report the average testing NDCG@3 of all testing task languages using lambda ranking method and single features.

% ----------------------------------------------------

% \subsection{NDCG evaluation metric}

%We evaluate the performance of the ranking method by the normalized discounted cumulative gain (NDCG) at the third rank position.

\subsection{Method Parameters and Baselines}

%\gn{We need to describe here which methods we're evaluating. This includes both the proposed \langrank method, and the baselines that use only single features. When introducing the baselines, try to link them back to methods that have been used in previous work.}

We use GBDT to train our \langrank models. For each \langrank model, we train an ensemble of 100 decision trees, each with 16 leaves. We use the LightGBM implementation \cite{Ke2017} of the LambdaRank algorithm in our training. In our experiments, we set $\gamma_{\max} = 10$, and evaluate the models by NDCG@3. The threshold of 3 was somewhat arbitrary, but based on our intuition that we would like to test whether \langrank can successfully recommend the best transfer language within a few tries, instead of testing its ability to accurately rank \emph{all} available transfer languages. The results in Table \ref{table:MainResults} report the average NDCG@3 across all cross-validation folds. For \langrank (all) we include all available features in our models, while for \langrank (dataset) and \langrank (ling) we include only the subsets of dataset-dependent and dataset-independent features, respectively.

We consider the following baseline methods:
\begin{itemize}
\item Using a single dataset-dependent feature: While dataset-dependent features have not typically been used as criteria for selecting transfer languages, they are a common feature in data selection methods for cross-domain transfer \cite{moore2010intelligent}. In view of this, we include selecting the transfer languages by sorting against each single one of $o_w$, $o_{sw}$, and $s_{tf}/s_{tk}$ in descending order, and sorting against $d_{ttr}$ in ascending order, as baseline methods.
\item Using a single linguistic distance feature: More common heuristic criteria of selection the transfer languages are choosing ones that have small phylogenetic distance to the task language \cite{dong15multitask, cotterell17crosslingualmorphology}. We therefore include selecting the transfer languages by sorting against each single one of $d_{gen}$, $d_{syn}$, $d_{fea}$, $d_{pho}$, $d_{inv}$, and $d_{geo}$ in ascending order as our baseline methods.
\end{itemize}

% ----------------------------------------------------
% ----------------------------------------------------

\section{Results and Analysis}
\label{sec:results}

\begin{table}[t!]
\begin{center}
\small
\begin{tabular}{ll|rrrr}
\midrule
\multicolumn{2}{c|}{\bf Method} & \multicolumn{1}{c}{\bf MT} & \multicolumn{1}{c}{\bf EL} & \multicolumn{1}{c}{\bf POS} & \multicolumn{1}{c}{\bf DEP}\\
\midrule
\parbox[t]{2mm}{\multirow{4}{*}{\rotatebox[origin=c]{90}{dataset}}}& word overlap $o_w$ & 28.6 & 30.7 & 13.4 & 52.3 \\
&subword overlap $o_{sw}$ & 29.2 & -- & -- & -- \\
& size ratio $s_{tf} / s_{tk}$ & 3.7 & 0.3 & 9.5 & 24.8 \\
& type-token ratio $d_{ttr}$ & 2.5  & -- & 7.4 & 6.4 \\
\midrule
\parbox[t]{2mm}{\multirow{6}{*}{\rotatebox[origin=c]{90}{ling. distance}}}&genetic $d_{gen}$ & 24.2 & 50.9 & 14.8 & 32.0  \\
& syntactic $d_{syn}$ & 14.8  & 46.4  & 4.1 & 22.9 \\
& featural $d_{fea}$ & 10.1 & 47.5 & 5.7 & 13.9 \\
& phonological $d_{pho}$ & 3.0  & 4.0  & 9.8 & 43.4 \\
&inventory $d_{inv}$ & 8.5  & 41.3  & 2.4 & 23.5 \\
& geographic $d_{geo}$ & 15.1 & 49.5  & 15.7 & 46.4 \\
\midrule
\multicolumn{2}{l|}{\langrank (all)} & 51.1& \textbf{63.0} & \textbf{28.9} & \textbf{65.0} \\
\multicolumn{2}{l|}{\langrank (dataset)} & \textbf{53.7} & 17.0 & 26.5 & \textbf{65.0} \\
\multicolumn{2}{l|}{\langrank (URIEL)} & 32.6 & 58.1 & 16.6 & 59.6  \\
\midrule
\end{tabular}
\end{center}
\vspace{-3mm}
\caption{\label{table:MainResults} Our \langrank model leads to higher average NDCG@3 over the baselines on all four tasks: machine translation (MT), entity linking (EL),  part-of-speech tagging (POS) and dependency parsing (DEP).} 
\vspace{-4mm}
\end{table}

\subsection{Main Results}
%\label{subsec:Effectiveness}
The performance of predicting transfer languages for the four NLP tasks using single-feature baselines and \langrank is shown in Table~\ref{table:MainResults}.
First, using \langrank with either all features or a subset of the features leads to substantially higher NDCG than using single-feature heuristics. Although some single-feature baselines manage to achieve high NDCG for some tasks, the predictions of \langrank consistently surpass the baselines on all tasks. In fact, for the MT and POS tagging tasks, the ranking quality of the best \langrank model is almost double that of the best single-feature baseline.
%The predictions made by the \langrank{} models that simultaneously consider all or subsets of features achieve substantially higher NDCG than the best predictions made based on any single features.
%This is true for \langrank using all available features, and is mostly true for \langrank using subsets of features.

Furthermore, using dataset-dependent features on top of the linguistic distance ones enhances the quality of the \langrank predictions. The best results for EL and POS tagging are obtained using all features, while for MT the best model is the one using dataset-only features. The best performance on DEP parsing is achieved with both settings. \langrank with only dataset features outperforms the linguistics-only \langrank on the MT and POS tagging tasks.
It is, however, severely lacking in the EL task, likely because EL datasets lack most dataset features as discussed in the previous section; the EL data only consists of pairs of corresponding entities and not complete sentences as in the case of the other tasks' datasets.

In addition, it is important to note that \langrank with only linguistic database information still outperforms all heuristic baselines on all tasks. This means that our model is potentially useful even before any resources for the language and task of interest have been collected, and could inform the data creation process.
%Second, we can see that if data has more statistical features, including those statistical features can greatly enhance the quality of predictions.
%We observe that on MT and POS tasks, the \langrank models using all or the subset of features related to the dataset statistics give better predictions, while on EL task the linguistic distance features are more effective.
%This might be caused by the fact that the EL datasets lack most dataset-statistic-related features, because that the data for EL consist of only pairs of corresponding entities, rather than complete sentences or paragraphs as in the case of MT and POS data.
%\gn{It is interesting to note that the linguistics distance features are still pretty good on their own, because the linguistic distance features are the ones we will have in a situation where we need to gather resources to perform transfer. I'd discuss this with at least one sentence.}

\begin{figure}[!t]
\centering
\includegraphics[width=\columnwidth]{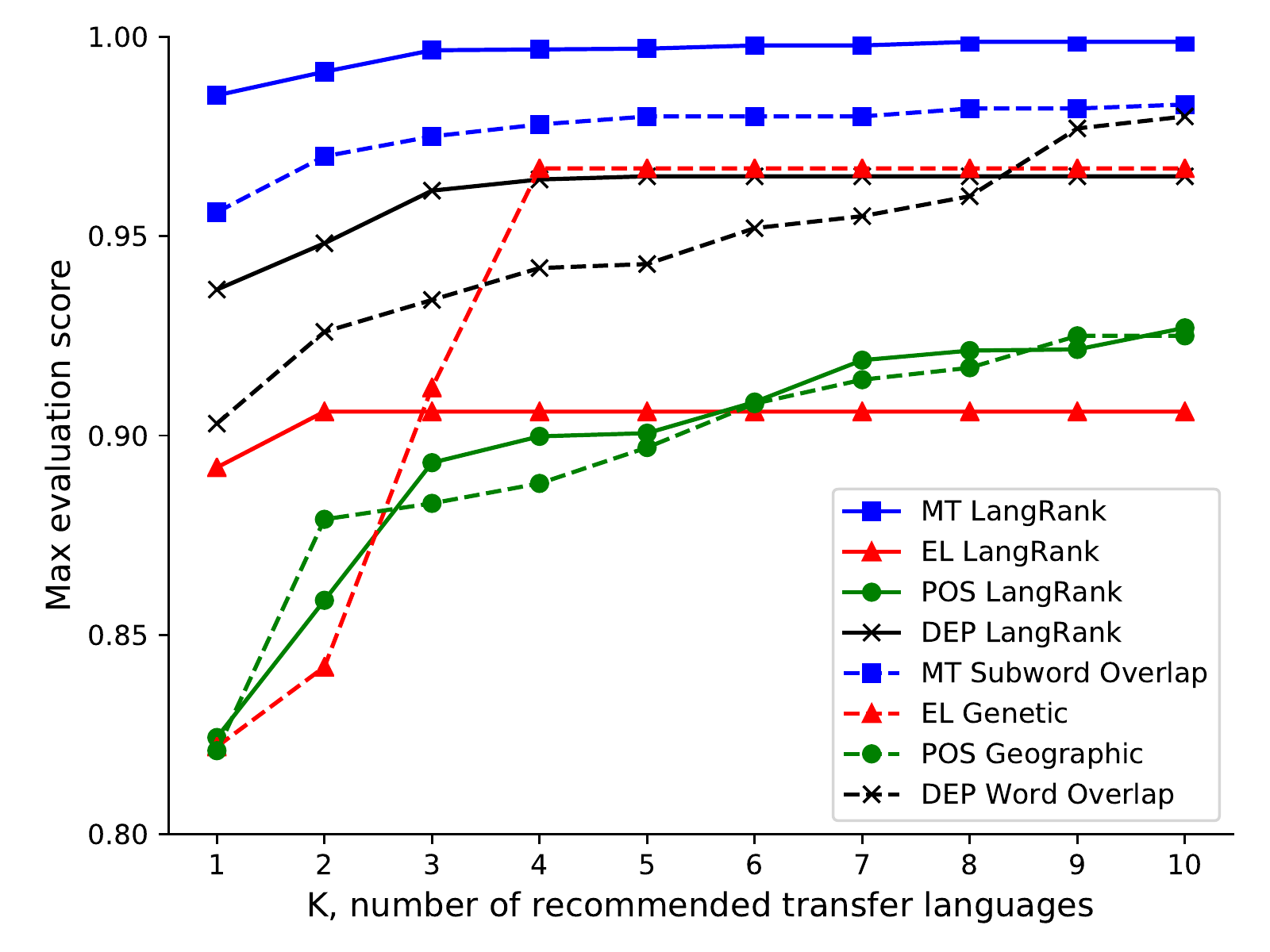}
\caption{The best evaluation score (BLEU for MT, accuracy for EL and POS, and LAS for DEP) attainable by trying out the top $K$ transfer languages recommended by the \langrank models and the single-feature baselines.}
\vspace{-5mm}
\label{fig:max_evaluation}
\end{figure}

\begin{table}[t!]
\centering
\resizebox{\columnwidth}{!}{
\small
\begin{tabular}{ccccc}
\toprule
\bf Task & \bf \textsc{Lang} & \bf Best & \bf Best & \bf True \\
\bf Lang & \bf \textsc{Rank} & \bf Dataset & \bf URIEL & \bf Best \\
\midrule
    &         & $o_w$   & $d_{fea}$ &  \\
 MT & tur (1) & tur (1) & ara (32) & tur (1) \\
aze & fas (3) & hrv (5) & fas (3) & kor (2) \\
    & hun (4) & ron (31) & sqi (22) & fas (3) \\
\iffalse
\midrule
    &         & $s_{tf} / s_{tg}$   & $d_{fea}$ &  \\
 MT & bul (1) & ara (9) & hun (2) & bul (1) \\
fin & hun (2) & heb (3) & bul (1) & hun (2) \\
    & vie (6) & rus (7) & rus (7) & pol (3) \\
\fi
\midrule
    &         & $o_w$   & $d_{geo}$ &  \\
 MT & hun (1) & vie (3) & mya (30) & hun (1) \\
ben & tur (2) & ita (20) & hin (27) & tur (2) \\
    & fas (4) & por (18) & mar (41) & vie (3) \\
\midrule
    &         & $o_w$   & $d_{inv}$ &  \\
 EL & amh (6) & amh (6) & pan (2) & hin (1) \\
tel  & orm (40) & swa (32) & hin (1) & pan (2) \\
    & msa (7) & jav (9) & ben (5) & mar (3) \\
\bottomrule
\end{tabular}
}
\caption{\label{table:LangExample} Examples of predicted top-3 transfer languages (and true ranks). The languages are denoted by the ISO 639-2 Language Codes. The first two task languages (aze, ben) are on the MT task, and the last one (tel) is on the EL task.}
\end{table}

% ----------------------------------------------------

% \subsection{Efficiency achieved by using the \langrank model} \gn{Commented out because it probably doesn't warrant a whole sub-section.}

Finally, from a potential user's point of view, a practical question is: \emph{If we train models on the top $K$ transfer languages suggested by the ranking model and pick the best one, how good is the best model expected to be?}
%``If I have resource to train $K$ transfer models using the top $K$ transfer languages recommended by the ranking model and pick the best one, how good the best model would be?''
If a user could obtain a good transfer model by trying out only a small number of transfer languages as suggested by our ranking model, the overhead of searching for a good transfer language is immensely reduced.

Figure \ref{fig:max_evaluation} compares the BLEU score (for MT), accuracy (for EL and POS) and LAS (for DEP) of the best transfer model attainable by using one of the top $K$ transfer languages recommended by \langrank (all) and by the best single feature baseline. We plot the ratio of the best score to that of the ground-truth best transfer model $c_{t,a_t^{*}}$, averaged over all task languages.
%\gn{Explain which systems are being compared.}
On the MT task, the best transfer models obtained by the suggestions of our \langrank (all) model constantly outperforms the models obtained from the best baseline.
%\gn{it's not ``any single feature'', it's specifically the ``genetic'' feature, right?}
On the POS tagging task, the best transfer models obtained by our ranking model are generally comparable to those using baseline suggestions.

% On the EL task, by trying less than 3 top transfer languages, \langrank{} gives better recommendations than the baselines, while if one is able to try more than 3 transfer languages, the baseline suggestions can lead to better transfer models.

We note that in the EL task, after looking beyond the top 3 \langrank predictions, the best baseline models on average seem to give more relevant transfer language suggestions than our \langrank models.
However, this is a case where averaging is possibly misleading. 
In fact, the \langrank model manages to select the correct top-1 language for 7 of the 9 task languages.
The other two languages (Telugu and Uyghur) do not have any typologically similar languages in the small training set, and hence the learned model fails to generalize to these languages.

In Table \ref{table:LangExample} we include a few representative examples of the top-3 transfer languages selected by \langrank and the baselines.%
\footnote{Detailed results are in the supplementary material.} %
In the first 
%two cases (aze and fin) 
case (aze) \langrank outperforms 
the already strong baselines by being able to consider both dataset and linguistic features, instead of considering them in isolation. 
In the 
%third 
second case (ben) where no baselines provide useful recommendations, \langrank still displays good performance; interestingly Turkish and Hungarian proved good transfer languages for a large number of task languages (perhaps to large data size and difficulty as tasks), and \langrank was able to learn to fall back to these when it found no good typological or dataset-driven matches otherwise -- behavior that would have be inconceivable without empirical discovery of transfer languages.
The final failure case (tel), as noted above, can be attributed to overfitting the small EL dataset, and may be remedied by either creating larger data or training \langrank jointly over multiple tasks.

%Since the best \langrank model relies more on typological features for the EL task, it fails to produce suitable transfer language suggestions in these two cases.\an{I reworded this, but I don't think it makes sense -- how is the Genetic distance baseline doing so much better, if the typological features are the ones that are supposed to be misleading the langrank model?}

%The poor performance may have two reasons. First, in all 9 task languages we test on, \langrank actually selects the true top-1 languages for 7 task languages, but fail on the other two. Since the performance is averaged over only 9 task languages, 2 failure cases make it apparently having nearly $1/4$ of the test cases fail. 
%Second, the task languages that \langrank fails, such as ``te'' (Table \ref{table:LangExample}), do not have typologically similar languages in the training set, so it is hard for the model to learn helpful information from the dataset features, and can only rely on the linguistic distance features. One can see indeed the best linguistic distance baseline predicts the top transfer languages much more successfully.

% ----------------------------------------------------

\subsection{Towards Better Educated Guesses for Choosing Transfer Languages}
\label{sec:educatedguesses}

\begin{figure}[t]
\centering
\includegraphics[width=\columnwidth]{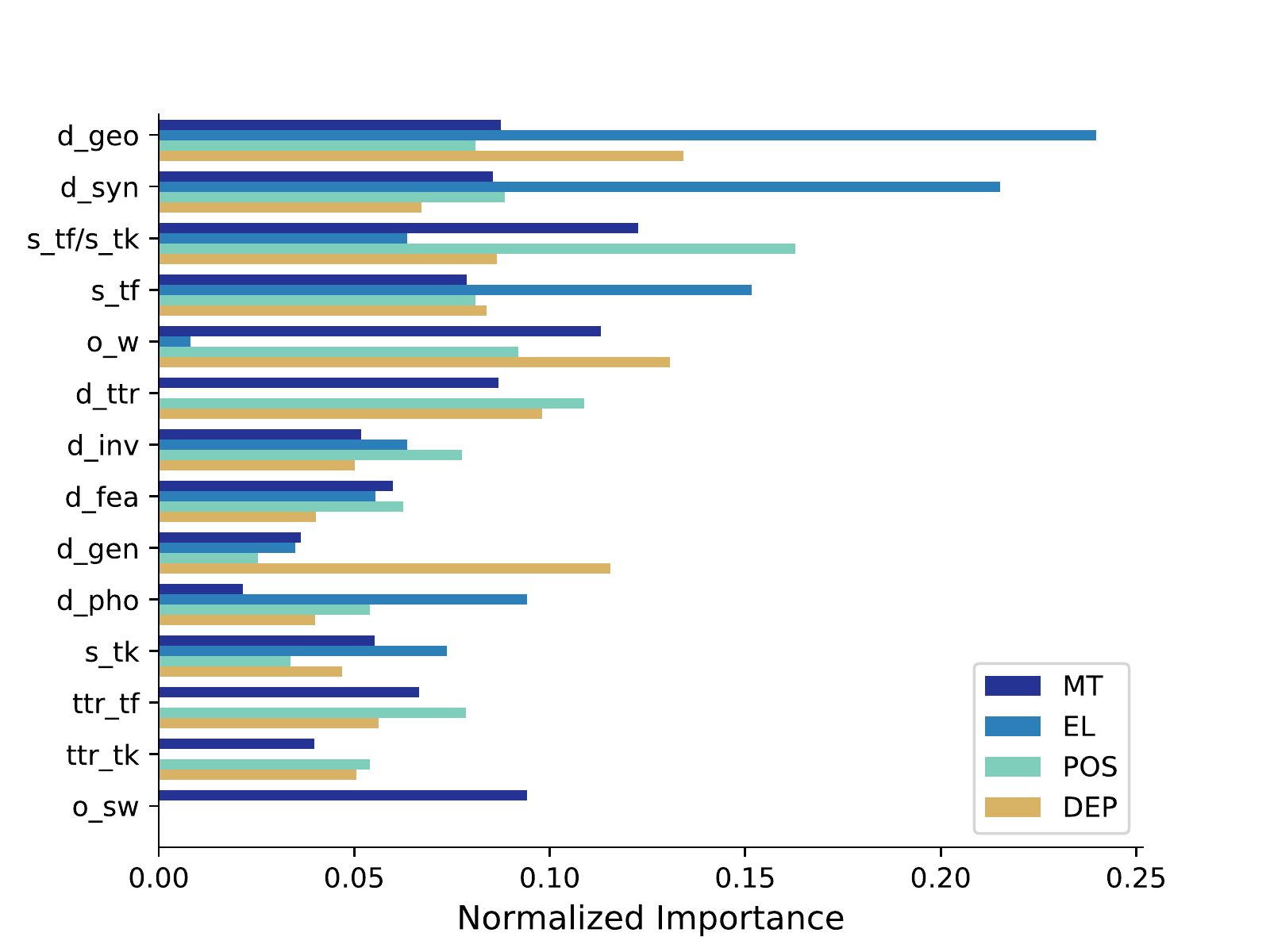}
\caption{Normalized feature importance for the MT, EL, POS and DEP tasks.}
% \vspace{-4mm}
\label{fig:Feature_importance}
\end{figure}

Our transfer language rankers are trained on a few languages for the particular tasks. It is possible that our models will not generalize well on a different set of languages or on other NLP tasks.
However, generating training data for ranking with exhaustive transfer experiments on a new task or set of languages will not always be feasible.
%For other NLP tasks or a very different set of languages, the existing models may not generalize well, and potentially the need to train a new ranking model for the new application arises, which requires generating the training data through exhaustive transfer experiments. \an{broke that sentence into smaller ones}
It could, therefore, be valuable to analyze the learned models and extract ``rules of thumb" that can be used as educated guesses in choosing transfer languages. They might still be ad-hoc, but they may prove superior to the intuition-based heuristic approaches used in previous work.
To elucidate how \langrank determines the best transfer languages for each task, Figure \ref{fig:Feature_importance} shows the feature importance for each of the NLP tasks.
The feature importance is defined as the number of times a feature is chosen to be the splitting feature in a node of the decision trees.
%Since a feature would only be chosen as the splitting feature if it leads to the largest information gain at a node, the above definition of feature importance is justified \gn{This sentence can be commented out if space is necessary.}.

% \gn{This is nice, but I can't help but feeling that maybe you're picking up idiosyncrasies of the dataset and explaining them in lots of detail despite the fact that they might not be 100\% generalizable. I wonder if there's anyway that we could focus on generalizable conclusions, such as the ones that were relatively consistent across datasets..} 

\begin{figure}[!t]
\centering
\small
\begin{tikzpicture}[every node/.style={inner sep=5pt}]

\node (start) [draw,rectangle,black] at (0,0) {$d_{gen} \leq 0.43$};
\node (out0) [draw,right=2cm of start] {output: 0};
\node (dec2) [draw,rectangle,black,below=.55cm of start] {$d_{syn} > 0.56$};
\node (out2) [draw,right=2cm of dec2] {output: 2};
\node (out3) [draw,below=.6cm of out2] {output: 3};
\node (dec3) [draw,rectangle,black,below=.5cm of dec2] {$\frac{s_{tf}}{s_{tk}} > 1.61$};
\node (out1) [draw,below=.4cm of out3] {output: 1};

\draw[->] (start.east) -- node[above,midway] {\small{yes}} (out0.west);
\draw[->] (start.south) -- node[auto,midway] {\small{no}} (dec2.north);

\draw[->] (dec2.east) -- node[above,midway] {\small{yes}} (out2.west);
\draw[->] (dec2.south) -- node[auto,midway] {\small{no}} (dec3.north);

\draw[->] (dec3.east) -- node[above,midway] {\small{yes}} (out3.west);
\draw[->] (dec3.south) |- +(0,-.67cm) -- node[above,midway] {\small{no}} (out1.west);
\end{tikzpicture}

\caption{An example of the decision tree learned in the machine translation task for Galician as task language.
%In each splitting node, $t$ denotes the value of the splitting threshold, and $g$ denotes the information gain.
}
% \vspace{-4mm}
\label{fig:Decision_tree}
\end{figure}
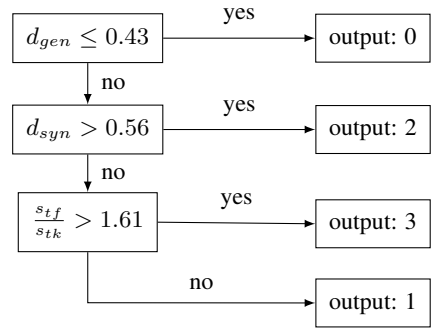

For the MT task, we find that dataset statistics features are more influential than the linguistic features, especially the dataset size ratio and the word overlap.
This indicates that a good transfer language for machine translation depends more on the dataset size of the transfer language corpus and its word and subword overlap with the task language corpus.
This is confirmed by results of the \langrank (dataset) model in Table \ref{table:MainResults}, which achieves the best performance by only using the subset of dataset statistics features.
At the same time, we note that the dataset size ratio and TTR distance, although of high importance among all features, when used alone result in very poor performance.
This phenomenon may be understood by looking at an example of a small decision tree in Figure~\ref{fig:Decision_tree}: a genetic distance of less than $0.4$ would produce a high ranking regardless of dataset size.
The dataset feature in this tree provides a smaller gain than two typological features, although it still informs the decision.
%is not always true that the dataset size alone entirely predicts the performance of a transfer langauge, but it helps to make more informative decision when considered together with other features.

For POS tagging, the two most important features are dataset size and the TTR distance. On the other hand, the lack of rich dataset-dependent features for the EL task leads to the geographic and syntactic distance being most influential.
There are several relatively important features for the DEP parsing task, with geographic and genetic distance standing out, as well as word overlap. These are features that also yield good scores on their own (see Table~\ref{table:MainResults}) but \langrank is able to combine them and achieve even better results.

\section{Related Work}

Cross-lingual transfer has been extensively used in several NLP tasks. In Section~\ref{sec:intro}, we provided a (non-exhaustive) list of examples that employ cross-lingual transfer across several tasks.
Other work has performed large-scale studies on the importance of appropriately selecting a transfer language, such as \citet{paul2009importance}, which performed an extensive search for a ``pivot language" in statistical MT, but without attempting to actually learn or predict which pivot language is best.

Typologically-informed models are another vein of research that is relevant to our work. The relationship between linguistic typology and statistical modeling has been studied by \newcite{gerz18typologyandlm} and
\citet{cotterell18hardtolanguagemodel}, with a focus on language modeling.
\citet{tsvetkov16polyglot} used typological information in the target language as additional input to their model for phonetic representation learning. \citet{ammar16manylanguages} and \citet{ahmad2018near} used similar ideas for dependency parsing, incorporating linguistically-informed vectors into their models.~\citet{ohoran-EtAl:2016:COLING} survey typological resources available and their utility in NLP tasks. 

\iffalse
Since typological information might not be readily available for every language, automatically producing or predicting such features is of interest. 
\citet{daumeiii07typology} built a Bayesian model in a large-scale attempt to predict implications between pairs of typological features, using over 2000 languages from the WALS database, and ~\citet{daumeiii:2009:NAACLHLT09} used statistical modeling to predict ``linguistic areas" and whether typology is shared within such areas. More recently, \citet{malaviya-neubig-littell:2017:EMNLP2017} trained a massively multilingual neural MT system that translates from over 1,000 languages to English to predict typological features, ~\citet{rabinovich-ordan-wintner:2017:Long} reconstruct phylogenetic trees from translated text and~\citet{bjerva_what_2019} examine the relationship between learned neural representations and language similarity. 
\fi

Although not for cross-lingual transfer, there has been prior work on data selection for training models.~\citet{tsvetkov_learning_2016} and~\citet{ruder_learning_2017} use Bayesian optimization for data selection.~\citet{van_der_wees_dynamic_2017} study the effect of data selection of neural machine translation, as well as propose a dynamic method to select relevant training data that improves translation performance.~\citet{plank-vannoord:2011:ACL-HLT2011} design a method to automatically select domain-relevant training data for parsing in English and Dutch. 

%\gn{TODO: This whole section has too high a ratio of Graham's/CMU's papers. We need to add more from other people}

%\paragraph{Cross-lingual Transfer}
%LM: \cite{tsvetkov16polyglot}
%MT: \cite{dong15multitask,johnson17googlemultilingual,Neubig2018}
%NER: \gn{TODO: add Mayhew cheap translation, JT's cross-lingual transfer} 

%Syntactic Analysis: \cite{tackstrom12crosslingualsyntax,ammar16manylanguages,cotterell17crosslingualmorphology,ahmad2018near} \sr{Parsing: Zeman and Resnik (2008) and Hwa, Resnik, Weinberg (2005)} 

%\paragraph{Incorporating Typology into Neural Models}
%\cite{tsvetkov16polyglot,ammar16manylanguages,ahmad2018near}
%TODO: summarize.

%\paragraph{Relationship Between Linguistic Features and Modeling}
%\cite{gerz18typologyandlm}
%\cite{cotterell18hardtolanguagemodel}

% ----------------------------------------------------
% ----------------------------------------------------

\section{Conclusion}

We formulate the task of selecting the optimal transfer languages for an NLP task as a ranking problem.
For machine translation, entity linking, part-of-speech tagging, and dependency parsing, we train ranking models to predict the most promising transfer languages to use given a task language.
We show that by taking multiple dataset statistics and language attributes into consideration, the learned ranking models recommend much better transfer languages than the ones suggested by considering only single language or dataset features.
Through analyzing the learned ranking models, we also gain some insights on the types of features that are most influential in selecting transfer languages for each of the NLP tasks, which may inform future \emph{ad hoc} selection even without using our method.

% In average, by training with no more than the top 3 transfer languages selected by our ranking models, the users can obtain better transfer models than they would do with the top 3 languages picked according to any single features.

%\gn{In the references, make sure everything that needs to be capitalized is capitalized. Also, make sure that everything that should have a conference/journal name does have one.} \yhl{Zirui and Pauline have checked.}

% ----------------------------------------------------
% ----------------------------------------------------

\section*{Acknowledgments}

This project was supported in part by NSF Award No. 1761548 ``Discovering and Demonstrating Linguistic Features for Language Documentation,'' and the Defense Advanced Research Projects Agency Information Innovation Office (I2O) Low Resource Languages for Emergent Incidents (LORELEI) program under Contract No. HR0011-15-C0114. The views and conclusions contained in this doc- ument are those of the authors and should not be interpreted as representing the official policies, either expressed or implied, of the U.S. Government. The U.S. Government is authorized to reproduce and distribute reprints for Government purposes notwithstanding any copyright notation here on.

% ----------------------------------------------------
% ----------------------------------------------------

\bibliography{MachineLearning}
\bibliographystyle{acl_natbib}

% ----------------------------------------------------
% ----------------------------------------------------
\iffalse
% These need to be in a separate pdf, if you wanted to provide an appendix --AA
\appendix

\section{Appendices}
\label{sec:appendix}

\input{suggested_lang_mt_1.tex}
\input{suggested_lang_mt_2.tex}
\input{suggested_lang_mt_3.tex}
\input{suggested_lang_mt_4.tex}

\input{suggested_lang_el.tex}

\input{suggested_lang_pos_1.tex}
\input{suggested_lang_pos_2.tex}
\fi

% ----------------------------------------------------
% ----------------------------------------------------

\end{document}